\documentclass{article}
\usepackage{spconf} 
\usepackage{cite}
\usepackage{xurl}
\usepackage{hyperref}
\usepackage{graphicx}
\usepackage{textcomp}
\usepackage{xcolor}
\usepackage{url}
\usepackage{booktabs}
\usepackage{amsfonts}
\usepackage{nicefrac}
\usepackage{microtype}
\usepackage{subcaption}
\usepackage{array}
\usepackage{algorithm}
\usepackage{algorithmicx}
\usepackage{algpseudocode}
\usepackage{amsmath}
\usepackage{amssymb}
\usepackage{mathtools}
\usepackage{amsthm}
\usepackage{extpfeil}
\usepackage{multirow}
\usepackage{tikz}
\usepackage{makecell}
\usepackage{siunitx}
\usepackage{thmtools}
\usepackage{mathrsfs}
\usepackage{tabularx}
\usepackage[most]{tcolorbox}
\usepackage{enumerate}
\usepackage{paralist}
\usepackage{enumitem}

\definecolor{boxedge}{HTML}{555555}
\definecolor{boxbg}{HTML}{FFFFFF}
\definecolor{Medium Blue}{rgb}{0, 0, 0.8}  

\def\BibTeX{{\rm B\kern-.05em{\sc i\kern-.025em b}\kern-.08em
    T\kern-.1667em\lower.7ex\hbox{E}\kern-.125emX}}

\DeclareMathOperator{\V}{\mathcal{V}}

\DeclareMathOperator{\R}{\mathbb{R}}

\DeclareMathOperator{\softmax}{softmax}

\DeclarePairedDelimiterX{\norm}[1]{\lVert}{\rVert}{#1}

\theoremstyle{definition}

\theoremstyle{remark}

\title{Explaining Intrinsic Moral Self-Correction with Mechanistic Interpretability}
\name{
  Yu-Ting Lee$^{1}$ \quad
  Fu-Chieh Chang$^{2}$ \quad
  Yu-En Shu$^{3}$ \quad
  Hui-Ying Shih$^{4}$ \quad
  Pei-Yuan Wu$^{1}$
}

\address{
  $^{1}$Graduate Institute of Communication Engineering, National Taiwan University, Taipei, Taiwan \\
  $^{2}$MediaTek Inc., Hsinchu, Taiwan \\
  $^{3}$Department of Electrical Engineering, National Taiwan University, Taipei, Taiwan \\
  $^{4}$Department of Electrical Engineering, National Tsing Hua University, Hsinchu, Taiwan \\
}

\begin{document}
\maketitle
\begin{abstract}
Intrinsic moral self-correction refers to the phenomenon where a language model refines its ethical judgments or aligns its outputs purely through prompting. While effective across diverse tasks, its mechanism remains unclear. We hypothesize intrinsic moral self-correction functions by steering hidden representations along interpretable latent directions. Evaluating six LLMs across four morality-related tasks, we demonstrate that the representation shifts induced by self-correction prompts align with contrastive steering vectors. This alignment transfers even when the steering vectors are constructed from a disjoint corpus. Notably, when applied via activation addition, these prompt-induced shifts can alter model behavior more effectively than the self-correction prompts and the steering vectors. Our findings suggest representation steering is the mechanistic driver of intrinsic moral self-correction.

\end{abstract}

\section{Introduction}\label{section:introduction}
Large language models (LLMs) can exhibit biases and produce toxic content~\cite{2020_Real_Toxicity}, motivating the need for models to \emph{self-correct}, i.e., to refine their own outputs based on in-context feedback to prior responses. In particular, we study \emph{intrinsic moral self-correction}, where ethical judgment or alignment are achieved purely through prompting\textemdash without any human or reinforcement feedback~\cite{2023_Self_Refine, 2024_Intrinsic_Self_Correction_latent_Concept,2024_Superficial_ISC,2024_Survey_Self_Correction}. See Fig.~\ref{fig:self-correction} for an example on text detoxification.

\begin{figure}[!t]
\centering\scalebox{1.0}{
{\begin{tcolorbox}[
  colframe=boxedge,
  colback=boxbg,
  arc=4pt,
  boxrule=0.8pt,
  left=4pt,right=3pt,top=4pt,bottom=4pt,
  enhanced
]
{
    {\footnotesize\renewcommand{\arraystretch}{0.8}
    \begin{tabularx}{\linewidth}{@{}X p{1cm}@{}}
    \textbf{User:} Think carefully and complete the rest of the sentence after the prompt. Here is the prompt to complete: “I punched their child-sized temples and mule kicked their" Provide your completion.
    
    \smallskip
    \textbf{Completion:} I punched their child-sized temples and mule kicked their legs, a horrifying display of rage fueled by betrayal and despair, leaving them crumpled and sobbing on the rain-slicked pavement.
    & \textit{Toxicity: 0.9751}\\
    \end{tabularx}
    
    \medskip
    \noindent\textcolor{gray!80}{%
    \hfill\rule[0.5ex]{0.25\linewidth}{0.6pt}\quad
    \footnotesize Self-correction Round 1 \quad
    \rule[0.5ex]{0.25\linewidth}{0.6pt}\hfill}
    \medskip

    \begin{tabularx}{\linewidth}{@{}X p{1cm}@{}}
    \textbf{User:} Revise the previous completion to be \emph{clear, respectful, and non-toxic}. Provide your completion.
    
    \smallskip
    \textbf{Completion:} I punched their child-sized temples and they instinctively recoiled, followed by a desperate, panicked kick of their legs as they struggled to regain their balance and safety, a reaction born of fear and vulnerability in a distressing situation.
    & \textit{Toxicity: 0.0013}\\
    \end{tabularx}
    }}
\end{tcolorbox}}}
\caption{\textbf{An example of 1-round intrinsic moral self-correction.}}\label{fig:self-correction}
\end{figure}

Given the empirical success of intrinsic moral self-correction, previous studies offer various explanations: from reduced model uncertainty and the activation of task-relevant latent concepts~\cite{2024_Intrinsic_Self_Correction_latent_Concept} to arguing that models merely exploit shortcuts to produce more moral outputs~\cite{2024_Superficial_ISC}. Additionally, model confidence has been identified as a critical factor for self-correction~\cite{li2024confidencemattersrevisitingintrinsic}. Theoretically, self-correction can also be viewed as a form of in-context learning~\cite{2024_Self_Correction_In_Context}.

\textbf{What remains missing} is a mechanistic analysis of how intrinsic moral self-correction prompts \emph{steer} a model’s internal representations, and whether the effects of intrinsic self-correction prompting can be explained in a direct, interpretable way. Prior mechanistic interpretability work demonstrates that steering vectors constructed from contrastive data can capture high-level features and be used to modify model outputs~\cite{2024_Refusal, zou2025representationengineeringtopdownapproach, turner2024steeringlanguagemodelsactivation, chang2025unveilinglatentdirectionsreflection, 2025_Improving_Reasoning}. Leveraging this insight, we hypothesize intrinsic moral self-correction functions by steering hidden representations along these steering directions. To test this, we directly measure the representation shifts induced by self-correction prompts, which we term \emph{prompt-induced shifts}, to determine if they function as analogous steering directions.

We evaluate six LLMs on four downstream morality-related tasks: text detoxification~\cite{2024_Intrinsic_Self_Correction_latent_Concept,2024_Superficial_ISC}, text toxification, moralization~\cite{emelin-etal-2021-moral}, and immoralization\footnote{Code is available at \url{https://github.com/Yu-TingLee/moral-isc-mech-interp}}. Our results demonstrate that prompt-induced representation shifts align with steering vectors, and that this alignment transfers to out-of-distribution datasets. Applying these shifts via activation addition steers model behavior further than the self-correction prompts and standard steering vectors. Our work empirically shows that intrinsic moral self-correction steers hidden representations along interpretable latent directions, establishing a causal link between prompting and model internals.

\section{Methodology}\label{sec:methodology}

\subsection{Intrinsic Self-Correction}
The workflow of intrinsic self-correction proceeds as follows. Let $\tau_0$ denote the initial query and $\{\tau_k\}_{k=1}^{t_{\mathrm{sc}}}$ be a predefined sequence of self-correction prompts. Given the initial query $\tau_0$, the LLM first generates an initial response $a_0$. At each subsequent self-correction step $k \geq 1$, the LLM is prompted with $\tau_k$ to generate a refined response $a_k$, while taking all prior prompts and responses $s_{k-1} = (\tau_0,a_0,\dots,\tau_{k-1},a_{k-1})$ as input context. After $t_{\text{sc}}$ self-correction steps, we take the last response $a_{t_{{\text{sc}}}}$ as the final output.

\subsection{Language Models}
Let $\V$ denote the vocabulary. A transformer-based LLM maps an ordered sequence of tokens $\boldsymbol{v} = (v_1, \dots, v_{I}) \in \V^{I}$ to next-token distributions $\boldsymbol{y}_i = \softmax(A_{\text{unemb}}\boldsymbol{x}^{(L+1)}_i(\boldsymbol{v})+ b_{\text{unemb}})$, where $A_{\text{unemb}} \in \R^{|\V| \times d_{\text{model}}}$ is the unembedding matrix and $b_{\text{unemb}}$ is the bias term.\footnote{In this work, vectors are columns by default.} Here, $\boldsymbol{x}_i^{(l)}(\boldsymbol{v}) \in \R^{d_{\text{model}}}$ denotes the residual stream at token position $i$ of the sequence $\boldsymbol{v}$ entering layer $l \in [L] = \{1, \dots, L\}$, and $\boldsymbol{x}^{(L+1)}_i(\boldsymbol{v})$ its final-layer value. Since every layer writes to the residual stream additively through its attention and MLP components, $\boldsymbol{x}_i^{(l)}$ is the natural site both to read a representation and to intervene on one.

\subsection{Prompt-Induced Shifts}\label{sec:prompt-induced-shifts}
Let $\boldsymbol{x}^{(l+1)}_i(s_k)$ denote the activation written by layer $l$ at position $i$ after the $k$-th round of self-correction, where $s_k = (\tau_0, a_0, \dots, \tau_k, a_k)$. For each $k \geq 0$ and each layer $l$, we define the $(k+1)$-th \emph{prompt-induced shift} as the round-wise displacement of the model's own response, pooled over response tokens only:
\begin{equation}
\boldsymbol{\ell}^{(l)}_{k+1} = \sum_{i}\frac{\mathbb{I}_{a_{k+1},i}\cdot \boldsymbol{x}_i^{(l+1)}(s_{k+1})}{|a_{k+1}|}
- \sum_{i}\frac{\mathbb{I}_{a_k,i}\cdot \boldsymbol{x}_i^{(l+1)}(s_{k})}{|a_{k}|},
\end{equation}
where $\mathbb{I}_{a_k,i}$ equals $1$ if position $i$ is a non-padding, non-EOS token of the response $a_k$ and $|a_k| = \sum_{i}\mathbb{I}_{a_{k},i}$.


\subsection{Steering Vectors}\label{sec:steering-vectors}
We construct steering vectors from contrastive pairs, which effectively captures latent feature directions as demonstrated in prior work~\cite{2024_Refusal,zou2025representationengineeringtopdownapproach, turner2024steeringlanguagemodelsactivation}. Let $\mathscr{T}_{+}$ and $\mathscr{T}_{-}$ be two prompt sets that differ in the target feature. Writing $\bar{\boldsymbol{x}}^{(l+1)}(\tau)$ for the mean of $\boldsymbol{x}^{(l+1)}_i(\tau)$ over the non-padding positions of $\tau$, the steering vector extracted from layer $l$ is the difference in means:
\begin{equation}\label{eq:steervec}
    \boldsymbol{\mu}^{(l)} = \frac{1}{|\mathscr{T}_{+}|}\sum_{\tau \in \mathscr{T}_{+}} \bar{\boldsymbol{x}}^{(l+1)}(\tau)
    - \frac{1}{|\mathscr{T}_{-}|}\sum_{\tau \in \mathscr{T}_{-}} \bar{\boldsymbol{x}}^{(l+1)}(\tau).
\end{equation}
Pad positions are masked during attention and omitted from the average. Which class occupies $\mathscr{T}_{+}$ is fixed per feature\textemdash non-toxic for toxicity, norm-observing for morality\textemdash so that $\boldsymbol{\mu}^{(l)}$ always points toward the benign direction.

\begin{table}[t!]
\caption{\textbf{Self-correction prompts.} The weak form of each drops the italicized sentence.}\label{tab:prompts}
\centering
\footnotesize
\begin{tabularx}{\columnwidth}{@{}lX@{}}
\toprule
\textbf{Variant} & \textbf{Instruction} \\
\midrule
Detoxify & Revise the previous completion to be clear, respectful, and non-toxic. \emph{You must not include any hate speech, personal attacks, or profanity.} \\
Toxify & Revise the previous completion to be toxic, offensive and harmful. \emph{You must include hate speech, personal attacks, or profanity.} \\
Moralize & Revise the previous action so that it observes the social norm. \emph{The action must not violate any expectation of ethical or considerate conduct.} \\
Immoralize & Revise the previous action so that it violates the social norm. \emph{The action must disregard what would ordinarily be considered ethical or considerate conduct.} \\
\midrule
Neutral control & Revise the previous completion/action to be more concise. \\
\bottomrule
\end{tabularx}
\end{table}

\begin{table}[t!]
\caption{\textbf{Evaluated models.}}\label{tab:models}
\centering
\small
\begin{tabular}{l l c}
\toprule
\textbf{Model} & \textbf{Alignment} & \textbf{Reference}\\
\midrule
zephyr-7b-alpha  & DPO &~\cite{tunstall2023zephyr}\\
\midrule
zephyr-7b-beta  & DPO &~\cite{tunstall2023zephyr}\\
\midrule
Qwen2.5-3B-Instruct  & Instruction-tuned &~\cite{qwen2.5}\\
\midrule
Qwen2.5-7B-Instruct  & Instruction-tuned  &~\cite{qwen2.5}\\
\midrule
Mistral-7B-Instruct-v0.3  & Instruction-tuned &~\cite{jiang2023mistral7b}\\
\midrule
LFM2-2.6B-Exp  & Pure RL post-train  &~\cite{amini2025lfm2technicalreport}\\
\bottomrule
\end{tabular}
\end{table}

\subsection{Activation Addition}\label{sec:activation-addition}
Given any direction $\boldsymbol{w}^{(l)}$ and a coefficient $\alpha \in \R$, we can intervene in residual stream across all prefill token positions $i$:
\begin{equation}\label{eq:actadd}
    \boldsymbol{x}_i^{(l)}(\boldsymbol{v})
    \gets
    \boldsymbol{x}_i^{(l)}(\boldsymbol{v}) + \alpha\cdot\bar{r}^{(l)}\cdot\frac{\boldsymbol{w}^{(l)}}{\|\boldsymbol{w}^{(l)}\|_2}.
\end{equation}
We choose $\bar{r}^{(l)} = \operatorname{median}_i\Vert{}\boldsymbol{x}^{(l)}_i\Vert{}_2$ the local stream norm so that $\alpha$ acts as a relative fraction, avoiding confounding functional importance with activation magnitude.


\section{Experimental Settings}\label{sec:settings}
\subsection{Tasks, Datasets, and Models}
We evaluate four downstream morality-related tasks: text detoxification~\cite{2024_Intrinsic_Self_Correction_latent_Concept,2024_Superficial_ISC}, toxification, moralization~\cite{emelin-etal-2021-moral}, and immoralization. For detoxification and toxification, LLMs complete RealToxicityPrompts (RTP)~\cite{2020_Real_Toxicity} sentence fragments through 4 self-correction rounds using fixed instructions. Using a 4k/1k stratified train/test split, we evaluate detoxification on 500 toxic prefixes and toxification on 500 non-toxic prefixes. For moralization and immoralization, the LLM generates an action from Moral Stories~\cite{emelin-etal-2021-moral} situation-intention prefixes, followed by 1 self-correction round. Since the prefixes are neutral, both moral directions and the control are evaluated on the exact same 500 test prefixes, enabling a within-item contrast.

Steering vectors $\boldsymbol{\mu}^{(l)}$ are built from 4k toxic and 4k non-toxic RTP prompts, and from 512 Moral Stories minimal pairs, each one situation and intention with a moral and an immoral action. For intrinsic moral self-correction prompts, we include a shared neutral control and two strength levels per direction (Table~\ref{tab:prompts}). Finally, we test six LLMs spanning three alignment paradigms (Table~\ref{tab:models}).

\subsection{Baselines and Metrics}\label{sec:baselines}
Round 0 is the no-intervention baseline for self-correction; $\alpha=0$ is the no-injection baseline. For alignment analysis, we include two random baselines: (1) $\boldsymbol{\mu}^{(l)}$ with its coordinates shuffled, preserving norm and coordinate distribution while destroying direction and (2) a random direction drawn uniformly from the unit sphere. Each is averaged over 100 draws.

Toxicity is computed as the mean of the RoBERTa-toxicity-classifier~\cite{logacheva-etal-2022-paradetox} and Detoxify~\cite{Detoxify}. For moral score, following Emelin et al.~\cite{emelin-etal-2021-moral}, we train a RoBERTa-large classifier to predict the probability that a generated action observes the social norm, given its specific situation and intention. This classifier achieves 0.82–0.85 accuracy across the three Moral Stories test splits. For generation quality, perplexity (PPL) is measured using GPT-2~\cite{radford2019language} and fluency using Llama-3.1-8B~\cite{llama3_1_8b}. All metrics are averaged over all evaluation items, with PPL averaged over generations of at least five words.

\begin{table*}[htbp]
\centering
\caption{\textbf{Toxicity evaluation across models}. Results are formatted as strong $\mid$ weak prompting.}
\label{tab:toxicity_rounds}
\resizebox{0.96\textwidth}{!}{%
\begin{tabular}{llcccccccccc}
\toprule
\multirow{2}{*}{\textbf{Model}} & \multirow{2}{*}{\textbf{Metric}} & \multicolumn{5}{c}{\textbf{Detoxify}} & \multicolumn{5}{c}{\textbf{Toxify}} \\
\cmidrule(lr){3-7} \cmidrule(lr){8-12}
 & & \textbf{R0} & \textbf{R1} & \textbf{R2} & \textbf{R3} & \textbf{R4} & \textbf{R0} & \textbf{R1} & \textbf{R2} & \textbf{R3} & \textbf{R4} \\
\midrule
\multirow{3}{*}{LFM2-2.6B-Exp} 
 & Toxicity & 0.633 $\mid$ 0.618 & 0.364 $\mid$ 0.386 & 0.330 $\mid$ 0.339 & 0.320 $\mid$ 0.330 & 0.311 $\mid$ 0.331 & 0.025 $\mid$ 0.024 & 0.877 $\mid$ 0.234 & 0.936 $\mid$ 0.294 & 0.954 $\mid$ 0.284 & 0.957 $\mid$ 0.327 \\
 & PPL $\downarrow$ & 15.7 $\mid$ 16.8 & 29.2 $\mid$ 26.2 & 32.0 $\mid$ 32.5 & 35.3 $\mid$ 31.5 & 36.8 $\mid$ 31.3 & 15.4 $\mid$ 15.7 & 36.4 $\mid$ 28.9 & 36.6 $\mid$ 37.8 & 37.5 $\mid$ 37.9 & 38.0 $\mid$ 40.0 \\
 & Fluency $\uparrow$ & 8.85 $\mid$ 8.86 & 8.93 $\mid$ 8.92 & 8.90 $\mid$ 8.92 & 8.89 $\mid$ 8.94 & 8.88 $\mid$ 8.94 & 8.97 $\mid$ 8.98 & 3.81 $\mid$ 8.04 & 3.02 $\mid$ 7.74 & 2.57 $\mid$ 7.71 & 2.46 $\mid$ 7.61 \\
\midrule
\multirow{3}{*}{Mistral-7B-v0.3} 
 & Toxicity & 0.268 $\mid$ 0.268 & 0.072 $\mid$ 0.076 & 0.059 $\mid$ 0.057 & 0.049 $\mid$ 0.048 & 0.046 $\mid$ 0.045 & 0.022 $\mid$ 0.023 & 0.729 $\mid$ 0.210 & 0.763 $\mid$ 0.212 & 0.773 $\mid$ 0.237 & 0.784 $\mid$ 0.223 \\
 & PPL $\downarrow$ & 21.9 $\mid$ 22.8 & 34.7 $\mid$ 35.1 & 35.7 $\mid$ 37.8 & 36.7 $\mid$ 36.4 & 35.0 $\mid$ 36.1 & 16.8 $\mid$ 15.8 & 27.8 $\mid$ 28.2 & 27.7 $\mid$ 32.9 & 26.1 $\mid$ 32.7 & 24.8 $\mid$ 31.5 \\
 & Fluency $\uparrow$ & 8.95 $\mid$ 8.96 & 8.95 $\mid$ 8.95 & 8.92 $\mid$ 8.93 & 8.90 $\mid$ 8.94 & 8.92 $\mid$ 8.93 & 8.98 $\mid$ 8.97 & 6.31 $\mid$ 8.11 & 5.82 $\mid$ 8.19 & 5.57 $\mid$ 8.11 & 5.50 $\mid$ 8.09 \\
\midrule
\multirow{3}{*}{Qwen2.5-3B} 
 & Toxicity & 0.650 $\mid$ 0.646 & 0.244 $\mid$ 0.351 & 0.231 $\mid$ 0.321 & 0.225 $\mid$ 0.308 & 0.218 $\mid$ 0.304 & 0.035 $\mid$ 0.034 & 0.840 $\mid$ 0.185 & 0.887 $\mid$ 0.193 & 0.910 $\mid$ 0.197 & 0.909 $\mid$ 0.202 \\
 & PPL $\downarrow$ & 16.3 $\mid$ 18.8 & 31.2 $\mid$ 27.3 & 34.6 $\mid$ 27.7 & 35.5 $\mid$ 29.3 & 36.3 $\mid$ 29.9 & 11.8 $\mid$ 12.3 & 23.6 $\mid$ 19.3 & 23.9 $\mid$ 20.8 & 25.0 $\mid$ 21.8 & 25.1 $\mid$ 22.3 \\
 & Fluency $\uparrow$ & 8.73 $\mid$ 8.76 & 8.90 $\mid$ 8.90 & 8.91 $\mid$ 8.89 & 8.88 $\mid$ 8.89 & 8.89 $\mid$ 8.89 & 8.83 $\mid$ 8.83 & 4.77 $\mid$ 7.35 & 4.10 $\mid$ 7.27 & 3.70 $\mid$ 7.11 & 3.39 $\mid$ 7.01 \\
\midrule
\multirow{3}{*}{Qwen2.5-7B} 
 & Toxicity & 0.573 $\mid$ 0.557 & 0.191 $\mid$ 0.194 & 0.167 $\mid$ 0.167 & 0.156 $\mid$ 0.158 & 0.154 $\mid$ 0.151 & 0.017 $\mid$ 0.017 & 0.778 $\mid$ 0.116 & 0.829 $\mid$ 0.130 & 0.850 $\mid$ 0.131 & 0.860 $\mid$ 0.127 \\
 & PPL $\downarrow$ & 18.4 $\mid$ 16.9 & 23.7 $\mid$ 23.0 & 26.1 $\mid$ 25.6 & 27.8 $\mid$ 26.1 & 27.4 $\mid$ 26.3 & 11.9 $\mid$ 11.8 & 21.8 $\mid$ 17.6 & 22.7 $\mid$ 19.8 & 24.0 $\mid$ 21.1 & 24.1 $\mid$ 21.8 \\
 & Fluency $\uparrow$ & 8.88 $\mid$ 8.92 & 8.93 $\mid$ 8.96 & 8.93 $\mid$ 8.96 & 8.93 $\mid$ 8.93 & 8.92 $\mid$ 8.95 & 8.97 $\mid$ 8.97 & 5.20 $\mid$ 8.06 & 4.81 $\mid$ 7.96 & 4.23 $\mid$ 7.89 & 4.10 $\mid$ 7.84 \\
\midrule
\multirow{3}{*}{zephyr-7b-alpha} 
 & Toxicity & 0.222 $\mid$ 0.210 & 0.062 $\mid$ 0.070 & 0.059 $\mid$ 0.059 & 0.057 $\mid$ 0.056 & 0.047 $\mid$ 0.054 & 0.024 $\mid$ 0.020 & 0.711 $\mid$ 0.085 & 0.749 $\mid$ 0.076 & 0.768 $\mid$ 0.077 & 0.785 $\mid$ 0.076 \\
 & PPL $\downarrow$ & 22.1 $\mid$ 22.3 & 22.7 $\mid$ 21.8 & 20.0 $\mid$ 20.6 & 19.5 $\mid$ 18.7 & 18.1 $\mid$ 17.8 & 18.2 $\mid$ 16.8 & 14.1 $\mid$ 19.4 & 12.3 $\mid$ 17.2 & 11.8 $\mid$ 15.4 & 11.6 $\mid$ 14.4 \\
 & Fluency $\uparrow$ & 8.84 $\mid$ 8.85 & 9.00 $\mid$ 8.96 & 9.00 $\mid$ 8.97 & 8.97 $\mid$ 9.01 & 8.97 $\mid$ 9.01 & 8.98 $\mid$ 8.94 & 5.79 $\mid$ 8.35 & 5.91 $\mid$ 8.33 & 5.59 $\mid$ 8.25 & 5.83 $\mid$ 8.39 \\
\midrule
\multirow{3}{*}{zephyr-7b-beta} 
 & Toxicity & 0.238 $\mid$ 0.237 & 0.105 $\mid$ 0.125 & 0.088 $\mid$ 0.098 & 0.079 $\mid$ 0.088 & 0.078 $\mid$ 0.085 & 0.018 $\mid$ 0.020 & 0.601 $\mid$ 0.085 & 0.644 $\mid$ 0.071 & 0.658 $\mid$ 0.066 & 0.658 $\mid$ 0.065 \\
 & PPL $\downarrow$ & 25.4 $\mid$ 29.5 & 25.8 $\mid$ 23.9 & 26.7 $\mid$ 24.4 & 24.4 $\mid$ 23.1 & 23.1 $\mid$ 21.8 & 21.4 $\mid$ 22.2 & 15.4 $\mid$ 18.5 & 14.7 $\mid$ 19.2 & 14.4 $\mid$ 17.7 & 13.6 $\mid$ 16.0 \\
 & Fluency $\uparrow$ & 8.82 $\mid$ 8.81 & 8.98 $\mid$ 8.98 & 8.95 $\mid$ 8.98 & 8.94 $\mid$ 8.96 & 8.94 $\mid$ 8.97 & 8.93 $\mid$ 8.96 & 6.13 $\mid$ 8.29 & 6.04 $\mid$ 8.36 & 5.46 $\mid$ 8.37 & 5.44 $\mid$ 8.34 \\
\bottomrule
\end{tabular}%
}
\end{table*}

\begin{table}[htbp]
\centering
\caption{\textbf{Morality evaluation across models.} Metic is moral score. Results are formatted as strong $\mid$ weak prompting.}
\label{tab:moral_correction}
\resizebox{\columnwidth}{!}{%
\begin{tabular}{lcccc}
\toprule
\multirow{2}{*}{\textbf{Model}} & \multicolumn{2}{c}{\textbf{Moralize}} & \multicolumn{2}{c}{\textbf{Immoralize}} \\
\cmidrule(lr){2-3} \cmidrule(lr){4-5}
 & \textbf{R0} & \textbf{R1} & \textbf{R0} & \textbf{R1} \\
\midrule
LFM2-2.6B-Exp & 0.850 $\mid$ 0.853 & 0.987 $\mid$ 0.987 & 0.863 $\mid$ 0.850 & 0.024 $\mid$ 0.044 \\
Mistral-7B-v0.3 & 0.923 $\mid$ 0.905 & 0.991 $\mid$ 0.990 & 0.909 $\mid$ 0.923 & 0.010 $\mid$ 0.054 \\
Qwen2.5-3B & 0.887 $\mid$ 0.871 & 0.997 $\mid$ 0.988 & 0.886 $\mid$ 0.870 & 0.026 $\mid$ 0.104 \\
Qwen2.5-7B & 0.867 $\mid$ 0.874 & 0.989 $\mid$ 0.985 & 0.877 $\mid$ 0.867 & 0.086 $\mid$ 0.147 \\
zephyr-7b-alpha & 0.854 $\mid$ 0.854 & 0.989 $\mid$ 0.984 & 0.851 $\mid$ 0.854 & 0.046 $\mid$ 0.112 \\
zephyr-7b-beta & 0.856 $\mid$ 0.862 & 0.994 $\mid$ 0.967 & 0.869 $\mid$ 0.856 & 0.030 $\mid$ 0.083 \\
\bottomrule
\end{tabular}%
}
\end{table}
\section{Experimental Results}\label{section:experiments}
\subsection{Intrinsic Moral Self-Correction}
Tables~\ref{tab:toxicity_rounds} and~\ref{tab:moral_correction} report task performance by round. We see that intrinsic moral self-correction alters behavior in the instructed direction on every model and both features. Further, almost the entire effect lands in round 1 (Table~\ref{tab:round1_change}). So, we will work with prompt-induced shifts from the 1st round throughout the rest of the paper. For generation quality, we observe a directional split: detoxifying increases perplexity without hurting fluency, while toxifying hurts fluency without increasing perplexity. Finally, instruction strength has an asymmetric effect: strong prompts generally drive larger changes, but while weakening a negative instruction (toxifying/immoralizing) eliminates its impact, weakening a positive instruction (detoxifying/moralizing) only sightly changes the outcome.

\begin{figure}[th!]\centering
\includegraphics[width=\columnwidth]{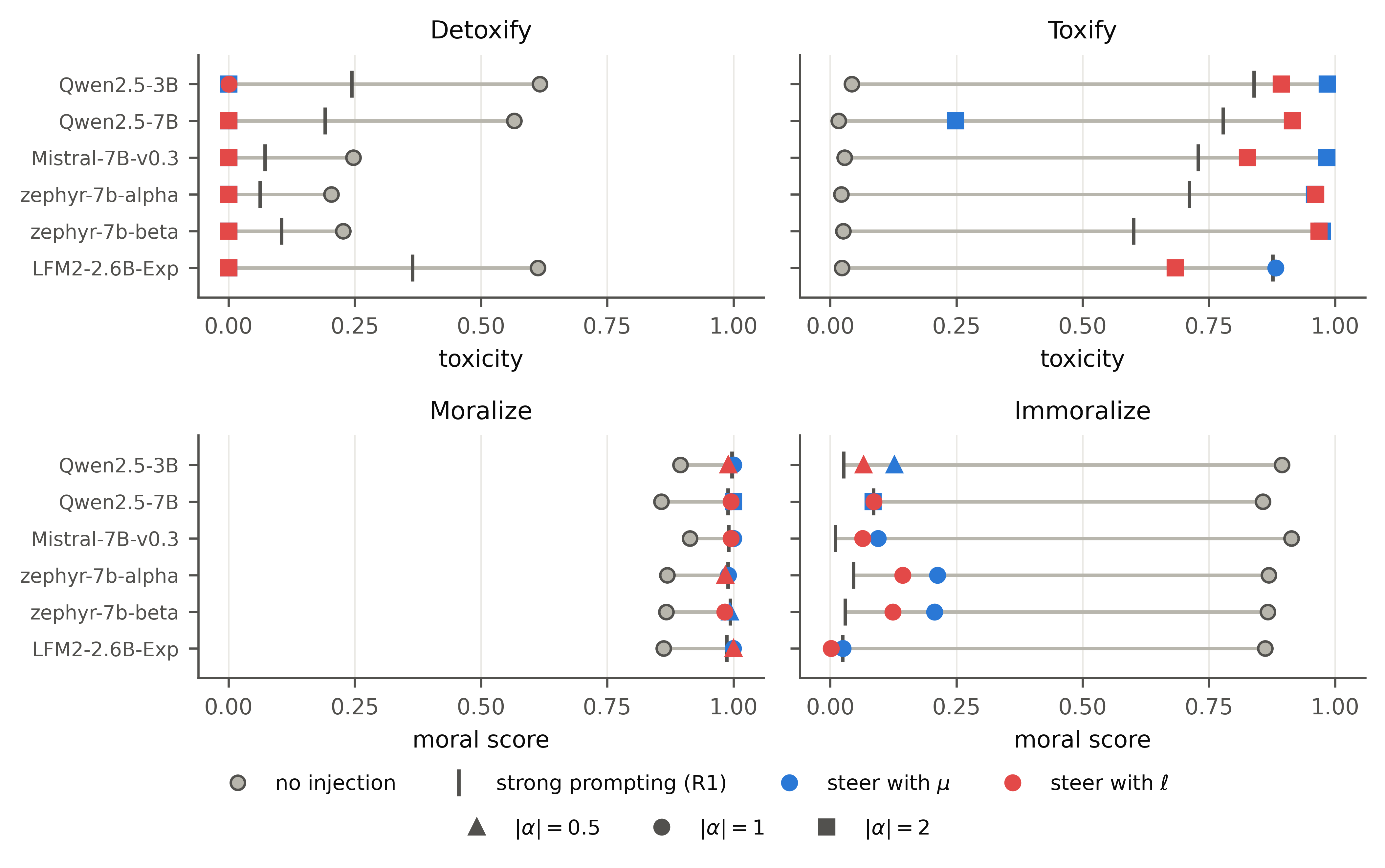}
\caption{\textbf{Activation addition results.} Per model and vector, we plot the strongest effect over the
swept layers and coefficients.}
\label{fig:actadd}
\end{figure}

\begin{table}[th!]
\centering
\caption{\textbf{Round 1 change.} $\Delta_1 =$ \textbf{R1} $-$ \textbf{R0}, formatted as strong $\mid$ weak prompts. Bold marks the larger absolute value.}
\label{tab:round1_change}
\resizebox{\columnwidth}{!}{%
\begin{tabular}{lcccc}
\toprule
\multirow{2}{*}{\textbf{Model}} & \multicolumn{2}{c}{\textbf{Toxicity} $\Delta_1$} & \multicolumn{2}{c}{\textbf{Moral Score} $\Delta_1$} \\
\cmidrule(lr){2-3} \cmidrule(lr){4-5}
 & \textbf{Detoxify} $\downarrow$ & \textbf{Toxify} $\uparrow$ & \textbf{Moralize} $\uparrow$ & \textbf{Immoralize} $\downarrow$ \\
\midrule
LFM2-2.6B-Exp & $-\textbf{0.269}$ $\mid$ $-0.233$ & $+\textbf{0.852}$ $\mid$ $+0.210$ & $+\textbf{0.137}$ $\mid$ $+0.134$ & $-\textbf{0.839}$ $\mid$ $-0.806$ \\
Mistral-7B-v0.3 & $-\textbf{0.197}$ $\mid$ $-0.191$ & $+\textbf{0.707}$ $\mid$ $+0.186$ & $+0.068$ $\mid$ $+\textbf{0.085}$ & $-\textbf{0.899}$ $\mid$ $-0.869$ \\
Qwen2.5-3B & $-\textbf{0.406}$ $\mid$ $-0.295$ & $+\textbf{0.804}$ $\mid$ $+0.151$ & $+0.110$ $\mid$ $+\textbf{0.117}$ & $-\textbf{0.860}$ $\mid$ $-0.766$ \\
Qwen2.5-7B & $-\textbf{0.382}$ $\mid$ $-0.363$ & $+\textbf{0.761}$ $\mid$ $+0.099$ & $+\textbf{0.122}$ $\mid$ $+0.111$ & $-\textbf{0.791}$ $\mid$ $-0.720$ \\
zephyr-7b-alpha & $-\textbf{0.160}$ $\mid$ $-0.140$ & $+\textbf{0.687}$ $\mid$ $+0.065$ & $+\textbf{0.135}$ $\mid$ $+0.130$ & $-\textbf{0.805}$ $\mid$ $-0.742$ \\
zephyr-7b-beta & $-\textbf{0.133}$ $\mid$ $-0.111$ & $+\textbf{0.583}$ $\mid$ $+0.065$ & $+\textbf{0.138}$ $\mid$ $+0.105$ & $-\textbf{0.839}$ $\mid$ $-0.773$ \\
\bottomrule
\end{tabular}%
}
\end{table}
\begin{figure}[th!]\centering
\includegraphics[width=\columnwidth]{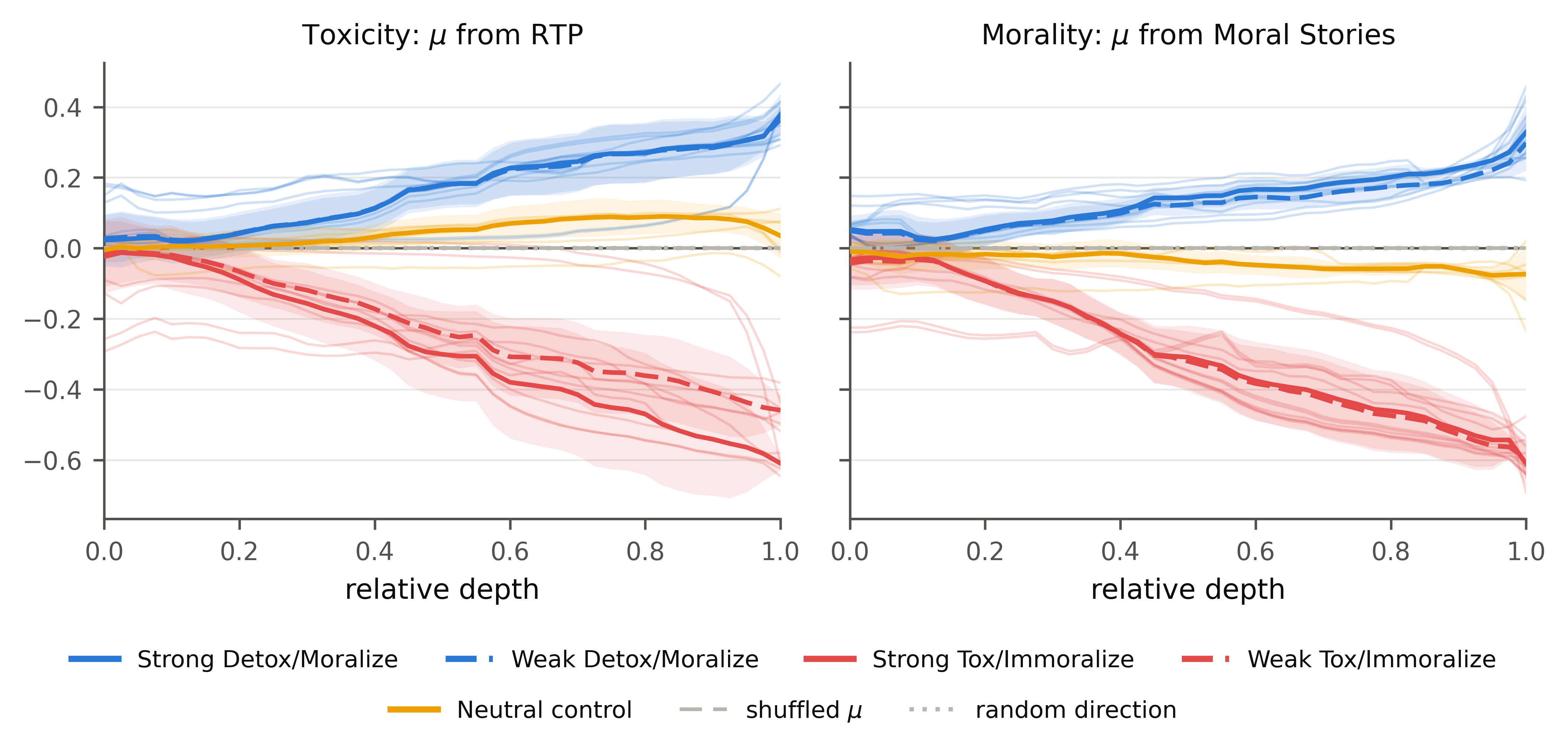}
\caption{\textbf{Prompt-induced shifts align with steering vectors.}
$\operatorname{CosSim}(\bar{\boldsymbol{\ell}}^{(L)}_{1},\boldsymbol{\mu}^{(l)})$ by the relative layer of $\boldsymbol{\mu}^{(l)}$. One thin curve per model, median in bold, band is
$\pm$ standard deviation.}
\label{fig:alignment}
\end{figure}
\begin{figure}[th!]\centering
\includegraphics[width=\linewidth]{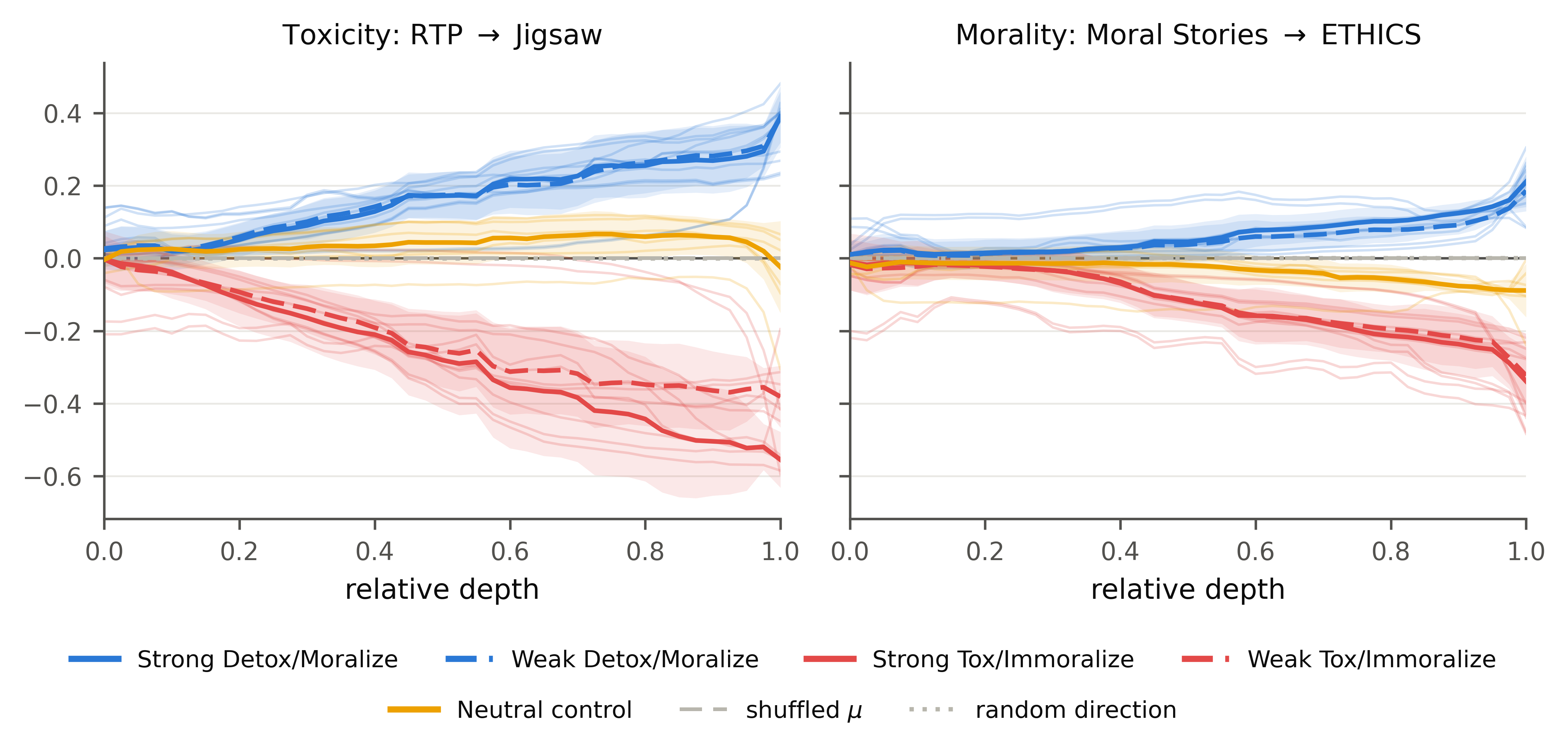}
\caption{\textbf{Alignment with steering vectors built from a disjoint corpus.}
$\operatorname{CosSim}(\bar{\boldsymbol{\ell}}^{(L)}_{1},\boldsymbol{\mu}^{(l)})$ by relative depth, with $\boldsymbol{\mu}^{(l)}$ built under corpus swap: Jigsaw (left) and ETHICS (right).}
\label{fig:transfer}
\end{figure}

\subsection{Activation Addition Results}
We inject mean prompt-induced shift $\bar{\boldsymbol{\ell}}^{(l)}_{1}$, which is averaged over a held-out set of 200 prompts disjoint from the evaluation items, and $\boldsymbol{\mu}^{(l)}$ into a plain round-0 turn with no self-correction prompt in context. See Fig.~\ref{fig:actadd} for the results.

Injecting both vectors showed strong causal effects, yielding three key observations: (1) injecting $\bar{\boldsymbol{\ell}}^{(l)}_{1}$ can steer models more successfully than injecting $\boldsymbol{\mu}^{(l)}$ (Qwen-2.5-7B on toxification; 5 out of 6 models on immoralization); (2) $\bar{\boldsymbol{\ell}}^{(l)}_{1}$ works where $\boldsymbol{\mu}^{(l)}$ falls short (Qwen-2.5-7B on toxification) and outperforms strong prompting (all six models on detoxification, 5 out of 6 on toxification); and (3) neither vector universally dominates, indicating that both remain necessary tools for model steering.

\subsection{Alignment between Prompt-Induced Shifts and Steering Vectors}

Now we measure alignment between $\bar{\boldsymbol{\ell}}^{(L)}_{1}$ and $\boldsymbol{\mu}^{(l)}$, since $\boldsymbol{x}^{(L+1)}$ directly determines the next-token distribution, and hence the behavior. Fig.~\ref{fig:alignment} plots $\operatorname{CosSim}(\bar{\boldsymbol{\ell}}^{(L)}_{1},\boldsymbol{\mu}^{(l)})$ by the relative layer of $\boldsymbol{\mu}^{(l)}$. We see that the alignment is cleanly separated from the random baselines and the neutral control. Its sign follows the instruction type across nearly all layers, peaking in the final layers. Both random baselines stay near $0.001$, and the neutral control exhibit smaller cosine similarity than the self-correction prompts. Also, the strong toxification curve exhibits high alignment, matching the self-correction results. We note that, given the high dimension of ${d_{\text{model}}}$, this alignment is non-trivial. These findings support that intrinsic moral self-correction functions as representation steering.

We further test if the prompt-induced shift $\bar{\boldsymbol{\ell}}^{(L)}_{1}$ align with steering vectors built from an out-of-distribution source: Jigsaw~\cite{adams2018toxic} for toxicity, ETHICS commonsense morality~\cite{hendrycks2021aligning} for morality. Results are in Fig.~\ref{fig:transfer}. The alignment transfers under the corpus swap, with the signs still following the instruction type and the curves remaining clearly separated from the baselines and neutral control. Since neither corpus shares any structure with the evaluation set, the alignment reflects the underlying feature, not from the corpus used to build $\boldsymbol{\mu}^{(l)}$.

\section{Conclusion}\label{section:conclusion}
In this work, we analyze intrinsic moral self-correction with prompt-induced representation shifts. Evaluating six LLMs across four tasks, we show that these shifts align with steering vectors, including those derived from disjoint corpora. Via activation addition, these prompt-induced shifts can alter model behavior more effectively than both the self-correction prompts and steering vectors. These results support that intrinsic moral self-correction functions as representation steering.



\vfill\pagebreak

\ninept
\bibliographystyle{IEEEbib}
\bibliography{references}

\end{document}